\documentclass[letterpaper]{article} % DO NOT CHANGE THIS
\usepackage{aaai2026}  % DO NOT CHANGE THIS
\usepackage{times}  % DO NOT CHANGE THIS
\usepackage{helvet}  % DO NOT CHANGE THIS
\usepackage{courier}  % DO NOT CHANGE THIS
\usepackage[hyphens]{url}  % DO NOT CHANGE THIS
\usepackage{graphicx} % DO NOT CHANGE THIS
\usepackage{natbib}  % DO NOT CHANGE THIS AND DO NOT ADD ANY OPTIONS TO IT
\usepackage{caption} % DO NOT CHANGE THIS AND DO NOT ADD ANY OPTIONS TO IT
\usepackage{algorithm}
\usepackage{algorithmic}
\usepackage{xcolor}
\usepackage{graphicx}
\usepackage{amsmath}
\usepackage{capt-of}
\usepackage{adjustbox}
\usepackage{cuted}
\usepackage{subfigure}
\usepackage{xcolor} % Required for color definitions
\usepackage{colortbl} % Required for colored cells
\usepackage{graphicx} % Required for \resizebox
\usepackage{booktabs}
\usepackage{color}
\usepackage[utf8]{inputenc}
\usepackage{array}
\usepackage{multirow}
\usepackage{amssymb} 
\usepackage{tabularx}
\usepackage{tikz}

\newcommand{\figref}[1]{Figure~\ref{#1}}
\newcommand{\tabref}[1]{Table~\ref{#1}}

\newcommand{\equref}[1]{Equation~(\ref{#1})}

\newcommand*\samethanks[1][\value{footnote}]{\footnotemark[#1]}
\newcommand{\tikzxmark}{%
\tikz[scale=0.23] {
    \draw[line width=0.7,line cap=round] (0,0) to [bend left=6] (1,1);
    \draw[line width=0.7,line cap=round] (0.2,0.95) to [bend right=3] (0.8,0.05);
}}
\newcommand{\tikzcmark}{%
\tikz[scale=0.23] {
    \draw[line width=0.7,line cap=round] (0.25,0) to [bend left=10] (1,1);
    \draw[line width=0.8,line cap=round] (0,0.35) to [bend right=1] (0.23,0);
}}
\usepackage{newfloat}
\usepackage{listings}
\DeclareCaptionStyle{ruled}{labelfont=normalfont,labelsep=colon,strut=off} % DO NOT CHANGE THIS
\floatstyle{ruled}
\newfloat{listing}{tb}{lst}{}
\floatname{listing}{Listing}

\title{Infinite-Story: A Training-Free Consistent Text-to-Image Generation}
\author{
    Jihun Park\thanks{Equal contribution}, Kyoungmin Lee\samethanks, Jongmin Gim\samethanks,  Hyeonseo Jo, Minseok Oh, Wonhyeok Choi, Kyumin Hwang, Jaeyeul Kim, Minwoo Choi, Sunghoon Im\thanks{Corresponding author}
}
\affiliations{

    DGIST, South Korea\\
    \{pjh2857, kyoungmin, jongmin4422, gustj0510, harrymark0, smu06117, kyumin, jykim94, subminu, sunghoonim\}@dgist.ac.kr

}

\usepackage{bibentry}
\begin{document}

\maketitle
% \vspace{-20pt}
\begin{strip} 
 % \vspace{-65pt}
    % \vspace{60pt}
    \centering
    \includegraphics[width=.75 \linewidth]{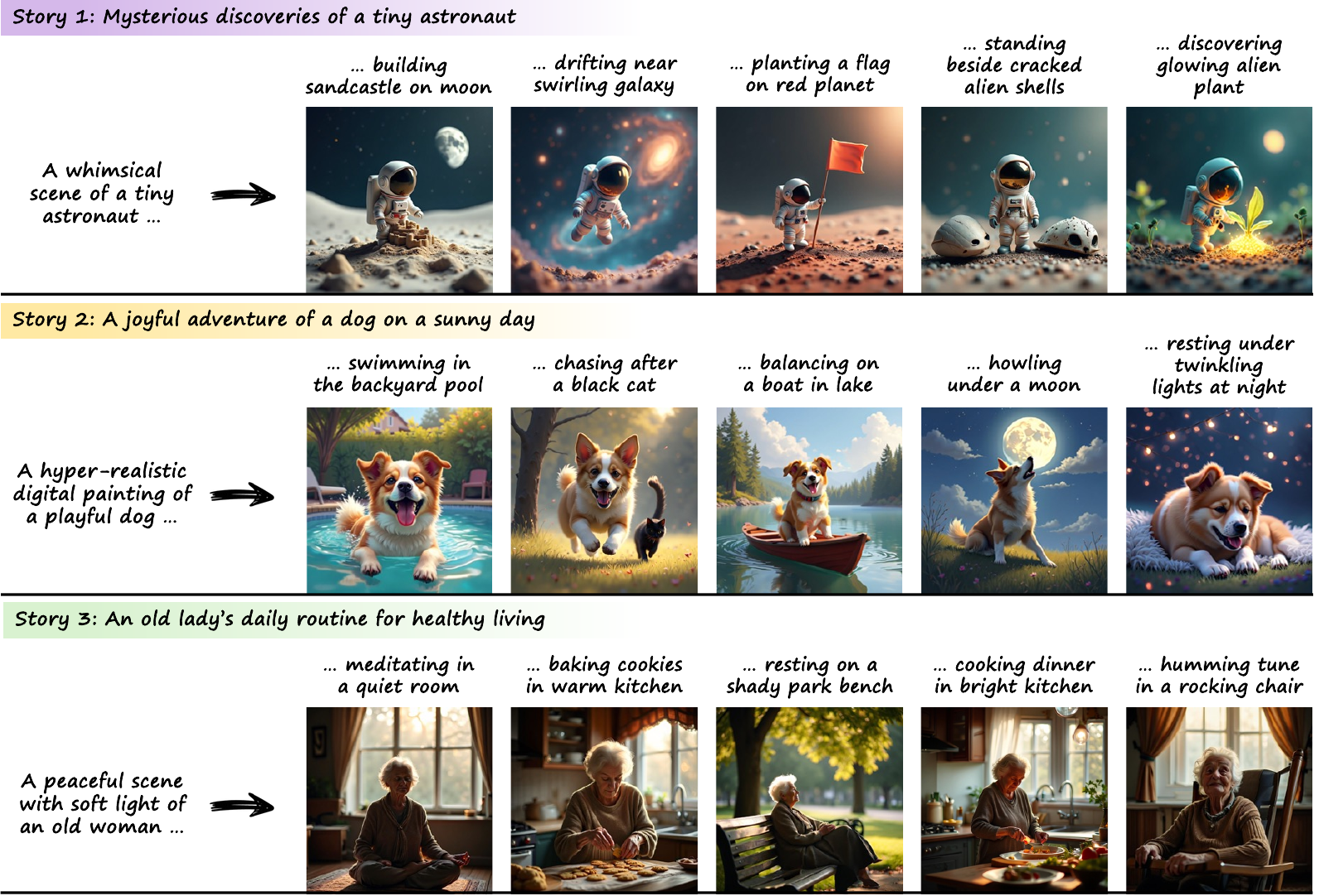}
    % \vspace{-60pt}
    \captionof{figure}{Results of Infinite-Story. Given text prompts with diverse expressions sharing identity prompts, our method generates image sequences with consistent subject identity and style.}

    \label{fig:teaser}
\end{strip}

\begin{abstract}
We present $\textit{Infinite-Story}$, a training-free framework for consistent text-to-image (T2I) generation tailored for multi-prompt storytelling scenarios. Built upon a scale-wise autoregressive model, our method addresses two key challenges in consistent T2I generation: identity inconsistency and style inconsistency. To overcome these issues, we introduce three complementary techniques: $\textit{Identity Prompt Replacement}$, which mitigates context bias in text encoders to align identity attributes across prompts; and a unified attention guidance mechanism comprising $\textit{Adaptive Style Injection}$ and $\textit{Synchronized Guidance Adaptation}$, which jointly enforce global style and identity appearance consistency while preserving prompt fidelity. Unlike prior diffusion-based approaches that require fine-tuning or suffer from slow inference, Infinite-Story operates entirely at test time, delivering high identity and style consistency across diverse prompts. Extensive experiments demonstrate that our method achieves state-of-the-art generation performance, while offering over 6$\times$ faster inference (1.72 seconds per image) than the existing fastest consistent T2I models, highlighting its effectiveness and practicality for real-world visual storytelling.

\end{abstract}

\begin{figure*}[t]
    \centering
    \includegraphics[width=0.75\linewidth]{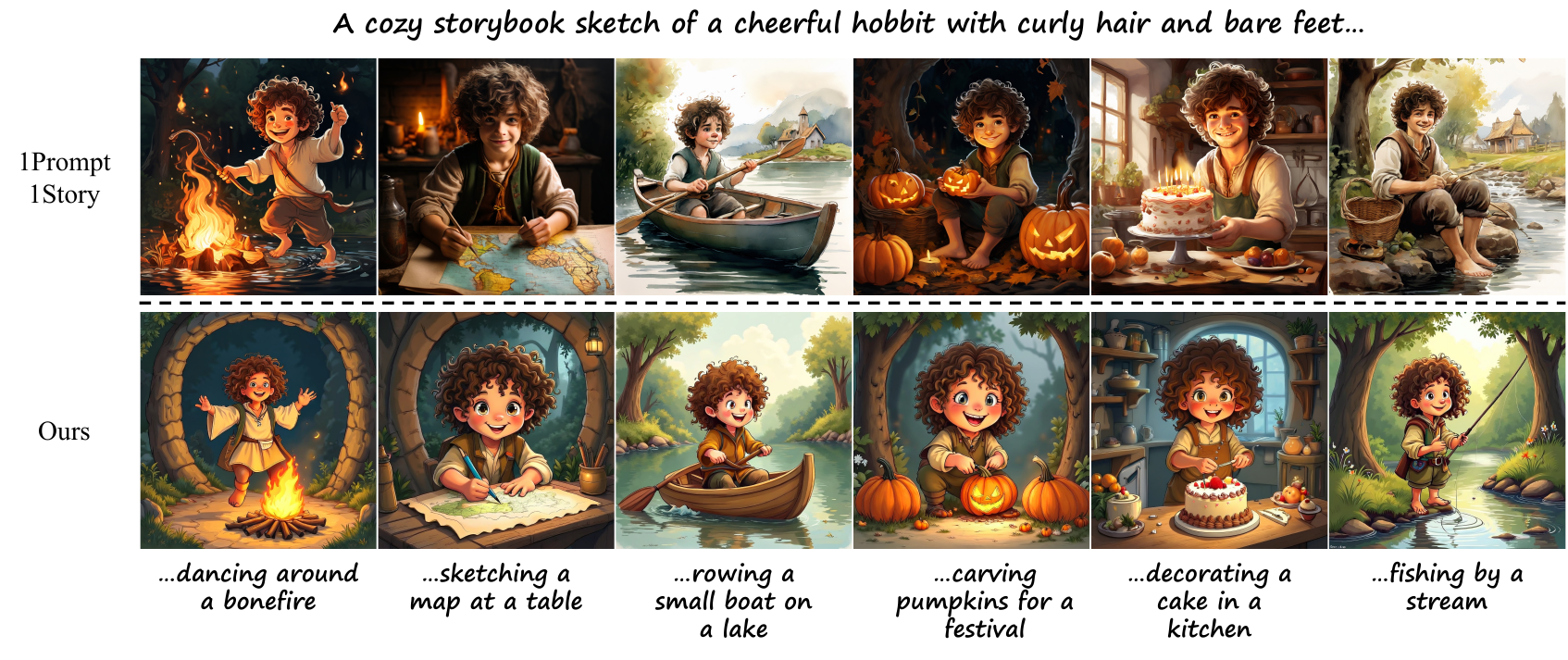}
    \caption{Qualitative comparison with 1Prompt1Story \cite{liu2025onepromptonestory}. While 1Prompt1Story maintains a consistent identity across the sequence, it fails to preserve visual style consistency, resulting in noticeable differences in rendering, background, and color tone across images. In contrast, our Infinite-Story achieves both identity and style consistency, producing a coherent visual narrative with uniform illustration style and subject appearance throughout the generated images.}
    \label{fig:intuition}
\end{figure*}

\section{Introduction}
\label{intro}
Large-scale diffusion-based Text-to-Image (T2I) generation models \cite{rombach2022high, ramesh2021zero, saharia2022photorealistic, podell2023sdxl, betker2023improving, flux2024} have demonstrated remarkable performance, establishing themselves as essential tools across a wide range of creative tasks, including design prototyping, content generation, visual communication, and advertising. However, the lack of consistency in generated images has posed limitations on user experience, particularly in scenarios that require coherence across multiple images, such as storytelling, character-driven content creation, comic strip generation, and sequential visual narratives.

To enforce consistency across generated images, various approaches have been proposed, including personalized image generation \cite{ruiz2023dreambooth, li2023blip, wei2023elite, ye2023ipadaptertextcompatibleimage}, 
style-aligned image generation \cite{park2025training, hertz2024style, sohn2023styledrop}, and consistent text-to-image generation \cite{avrahami2024chosen, liu2025onepromptonestory, tewel2024training, wang2024oneactor, zhou2024storydiffusion}. While consistent text-to-image generation is particularly foundational for visual storytelling tasks, prior works have largely focused on maintaining identity consistency across scenes. However, they often overlook style consistency between generated image sets, which is crucial for producing visually coherent narratives that span multiple scenes, as illustrated in \figref{fig:intuition}-(Top). In addition, most consistent text-to-image generation methods are based on diffusion models, which—even without fine-tuning—typically require over 10 seconds per image during inference. This surpasses the threshold at which users begin to lose focus during interactive sessions, according to Nielsen’s usability guidelines \cite{nielsen1994usability}.

Recently, scale-wise autoregressive models~\cite{tian2024visual, voronov2024swittidesigningscalewisetransformers, han2024infinity} have emerged as a promising alternative, offering faster inference by adopting a next-scale prediction paradigm. These models achieve competitive image quality while significantly improving inference speed compared to both traditional autoregressive~\cite{van2017neural, esser2021taming, chang2022maskgit, chang2023muse} and diffusion-based models~\cite{podell2023sdxl, flux2024}. While they effectively mitigate the latency issues inherent in diffusion approaches, scale-wise models continue to face challenges in ensuring consistency across generated images, such as identity inconsistency, style inconsistency, and a combination of both.

To address these challenges, we introduce \textit{Infinite-Story}, a training-free framework for consistent text-to-image generation built on a scale-wise autoregressive model~\cite{han2024infinity}, without modifying the architecture or requiring additional training. 
Our approach generates a set of images that retain consistency in both identity and style across varying prompts by designating one image in each batch as a reference and propagating its identity and style to guide the remaining samples.

\begin{figure}[t]
    \centering
    \includegraphics[width=.8\linewidth]{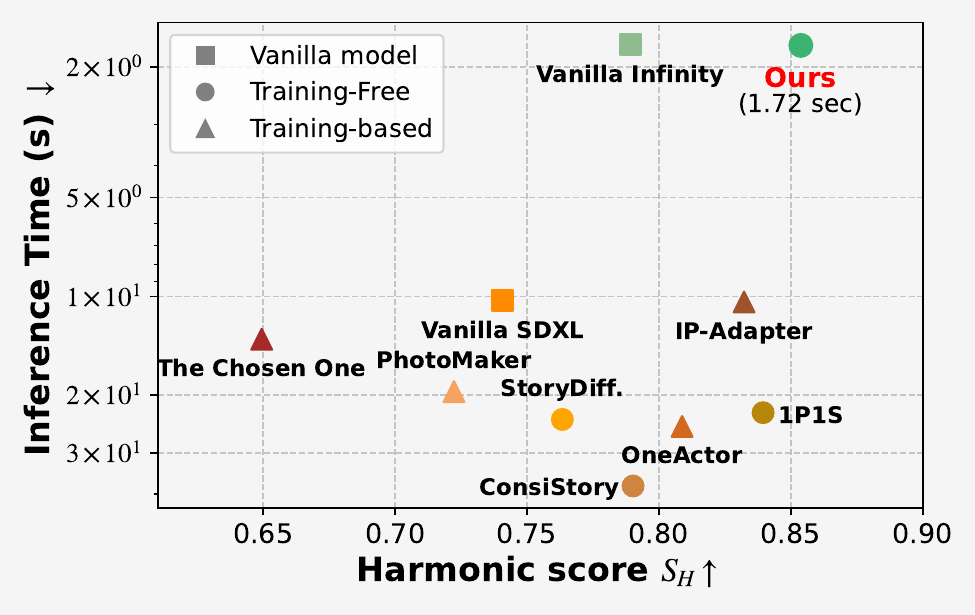}
    \caption{Comparison of inference time and harmonic score $S_H$ between our method and state-of-the-art identity-consistent text-to-image generation models.}
    \label{fig:time_performance}
\end{figure}

To this end, we propose three lightweight yet effective techniques: \textit{Identity Prompt Replacement}, which mitigates the context bias of text encoders to align identity-related attributes across prompts. Also, we propose a unified attention guidance that consists of \textit{Adaptive Style Injection} and \textit{Synchronized Guidance Adaptation}, enhancing both identity appearance and global visual style consistency via reference feature injection into early-stage self-attention layers, while ensuring prompt fidelity through synchronized adaptation across conditional and unconditional branches. These techniques are seamlessly integrated into the inference pipeline without any need for additional fine-tuning or training. By combining these components, Infinite-Story achieves state-of-the-art generation quality, as illustrated in \figref{fig:teaser} and \figref{fig:intuition}-(Bottom). It outperforms existing methods in both quantitative and qualitative evaluations, while also offering up to 6$\times$ faster inference time (1.72 seconds per image) than the fastest diffusion-based consistent T2I models, as shown in \figref{fig:time_performance}.

In summary, our primary contributions include:
\begin{itemize}
    \item We present \textit{Infinite-Story}, the first training-free, scale-wise autoregressive framework for consistent text-to-image generation.
    \item We introduce an \textit{Identity Prompt Replacement} technique that aligns identity attributes across prompts by unifying identity prompt embeddings.
    \item We propose a unified attention guidance approach that combines \textit{Adaptive Style Injection} and \textit{Synchronized Guidance Adaptation} to achieve consistent overall style and identity appearance while preserving prompt fidelity.

\end{itemize}

\section{Related work}
\label{related}
\subsection{Text-to-image generation}

Large-scale image-text datasets \cite{changpinyo2021conceptual, lin2014microsoft, schuhmann2022laion, kakaobrain2022coyo-700m} have enabled conditional image synthesis by bridging language and vision. This has spurred the development of powerful Text-to-Image (T2I) models—diffusion-based \cite{ramesh2021zero, rombach2022high, saharia2022photorealistic, podell2023sdxl,flux2024}, GAN-based \cite{kang2023scaling}, and autoregressive (AR)-based \cite{chang2023muse, han2024infinity, tang2024hartefficientvisualgeneration}—capable of producing high-quality images from text prompts. Diffusion models dominate with strong synthesis quality, supporting tasks like image editing \cite{brooks2023instructpix2pix, hertz2023delta, wang2023imagen} and translation \cite{tumanyan2023plug, parmar2023zero}, but suffer from slow inference. AR models have advanced from next-token prediction \cite{van2017neural, esser2021taming} to faster masked token generation \cite{chang2022maskgit, chang2023muse, kondratyuk2023videopoet}, with next-scale prediction \cite{tian2024visual} improving efficiency further \cite{han2024infinity, tang2024hartefficientvisualgeneration, voronov2024swittidesigningscalewisetransformers}. Nonetheless, T2I models still struggle with maintaining subject identity consistency across images, limiting their applicability in areas like storytelling, content creation, and branding.

\begin{figure*}[t]
    \centering
    \includegraphics[width=.9\linewidth]{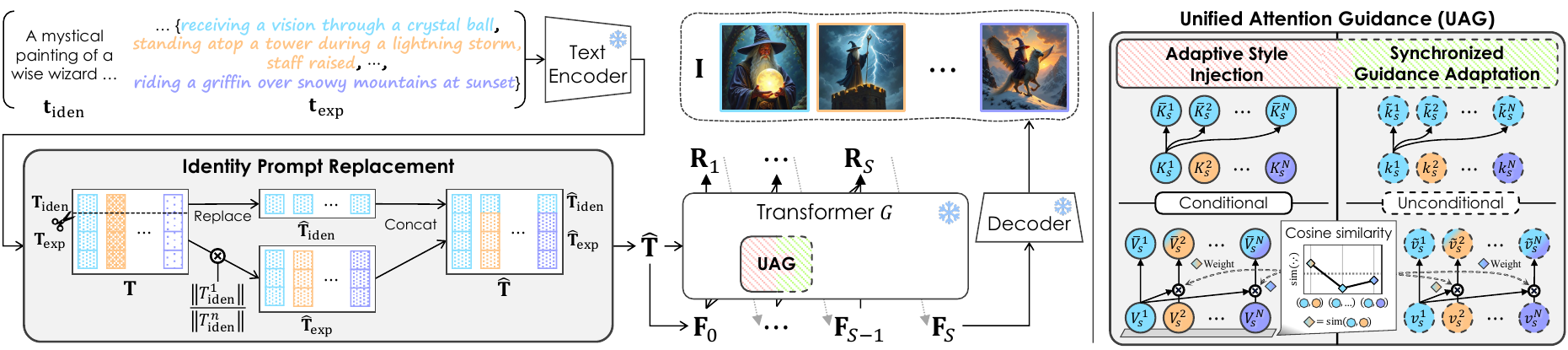}
    \caption{\textbf{Overall pipeline of our method.} The text encoder $E_T$ \cite{chung2022scalinginstructionfinetunedlanguagemodels} processes a set of text prompts $\mathbf{t}$, producing contextual embeddings $\mathbf{T}$ that condition the transformer. \textit{Identity Prompt Replacement} is applied to $\mathbf{T}$ before generation to ensure consistent identity representation across prompts. During generation, Unified Attention Guidance (UAG), which consists of \textit{Adaptive Style Injection} and \textit{Synchronized Guidance Adaptation}, is applied to early-stage self-attention layers to achieve consistent identity appearance and overall style alignment while preserving prompt fidelity. The transformer autoregressively produces residual feature maps, which are decoded into final images $\mathbf{I}$ via the image decoder.}
    \label{fig:overall}
\end{figure*}

\subsection{Personalized image generation}
Personalized image generation enables scene exploration with user-specific features. Existing methods are broadly categorized into subject-driven and style-driven approaches. Subject-driven methods \cite{li2023blip, gal2022image, wei2023elite, ye2023ipadaptertextcompatibleimage, ruiz2023dreambooth} typically fine-tune or adapt pre-trained encoders to inject concept embeddings from reference images, but often require external datasets, limiting generality. Recent works address this with parameter-efficient fine-tuning by updating limited model components like attention layers \cite{nam2024dreammatcher, kumari2023multi}. Style-driven methods instead focus on visual consistency by optimizing style features via LoRA-based tuning \cite{frenkel2024implicit, shah2024ziplora, sohn2023styledrop, hu2022lora, ryu2023low} or by adapting attention for stylistic coherence \cite{hertz2024style, park2025training}. Despite their strengths, most methods rely on diffusion models, which are slow and unsuitable for interactive use.

\subsection{Consistent text-to-image generation}
Consistent text-to-image generation, which aims to preserve identity across multiple images, has become a key focus within personalized image generation. Recent studies \cite{kumari2023multi, li2024photomaker, zhou2024storydiffusion, tewel2024training} show that adjusting attention layer weights effectively modulates identity. Other approaches \cite{mou2023t2iadapterlearningadaptersdig, zhang2023addingconditionalcontroltexttoimage} incorporate structured control to aid identity preservation. Foundational works \cite{radford2021learning, vaswani2017attention, devlin2019bertpretrainingdeepbidirectional, chen2025contrastivelocalizedlanguageimagepretraining, raffel2023exploringlimitstransferlearning} highlight the linguistic strength of transformer-based text encoders, while enhanced textual conditioning \cite{hertz2022prompt, gal2022image} further improves identity consistency. Building on this, \cite{liu2025onepromptonestory} leverage prompt embedding variations to maintain coherent identities across images. Inspired by these insights, we introduce a training-free consistent text-to-image generation method through the manipulation of prompt embeddings and attention mechanisms.

\section{Method}
\label{method}

\subsection{Overall pipeline}

In this paper, we aim to generate $N$ multiple images $\mathbf{I} = \{I^n\}_{n=1}^N$ from corresponding text prompts $\mathbf{t} = \{t^n\}_{n=1}^N$, each composed of the same identity prompts \(\mathbf{t}_{\text{iden}} = \{t_{\text{iden}}^n\}_{n=1}^N\) and distinct expression prompts  \(\mathbf{t}_{\text{exp}} = \{t_{\text{exp}}^n\}_{n=1}^N\), with the objective of maintaining consistent identity and overall style. All prompts are concatenated and processed in parallel as a batch. 

Our method is based on the Infinity architecture \cite{han2024infinity}, which employs a next-scale prediction scheme \cite{tian2024visual}. The model consists of a pre-trained text encoder $E_T$ employing Flan-T5 \cite{chung2022scalinginstructionfinetunedlanguagemodels}, a transformer $G$ that autoregressively predicts quantized residual $s$-th feature maps $\mathbf{R}_s$ over steps $\mathbf{S} = \{1,2,...,S\}$, and a decoder $D$ that reconstructs images from the final feature maps:
\begin{equation}
\begin{gathered}
    \mathbf{I} = D(\mathbf{F}_S), \quad
    \mathbf{F}_s = \sum_{i=1}^{s} \text{up}_{H\times W}(\mathbf{R}_i),~\mathbf{R}_s \in \mathbb{R}^{N \times h_s \times w_s}, \\
    \mathbf{R}_{s} = G(\mathbf{F}_{s-1}, \textbf{T}), \quad
    % =Attn_{cross}( Attn_{self}(\mathbf{Q}_{s-1}, \mathbf{K}_{s-1}, \mathbf{V}_{s-1}), \mathbf{K}_{t}, \mathbf{V}_{t},\mathbf{T}, \tau), \\
    \mathbf{T} = E_T(\mathbf{t}) = \left\{ \left( T_{\text{iden}}^n,\, T_{\text{exp}}^n \right) \right\}_{n=1}^N, 
\end{gathered}
\end{equation}
where $h_s$, $w_s$ are the spatial sizes at step $s\in\mathbf{S}$, $\text{up}_{H\times W} (\cdot)$ denotes a bilinear upsampling function to upsample to $H \times W$ size, and $\mathbf{T}$ denotes the encoded identity and expression features. The initial feature map $\mathbf{F}_0$ is initialized from $\mathbf{T}$.

As shown in \figref{fig:overall}, \textit{Identity Prompt Replacement} is first applied to ensure consistent identity attributes. During generation, both \textit{Adaptive Style Injection} and \textit{Synchronized Guidance Adaptation} are applied to self-attention layers in early generation steps $\mathbf{S}_{\text{early}}$, promoting consistent identity appearance and global style across all generated images.

\begin{figure}[t]
    \vspace{-10pt}
    \centering
    \includegraphics[width=.8\linewidth]{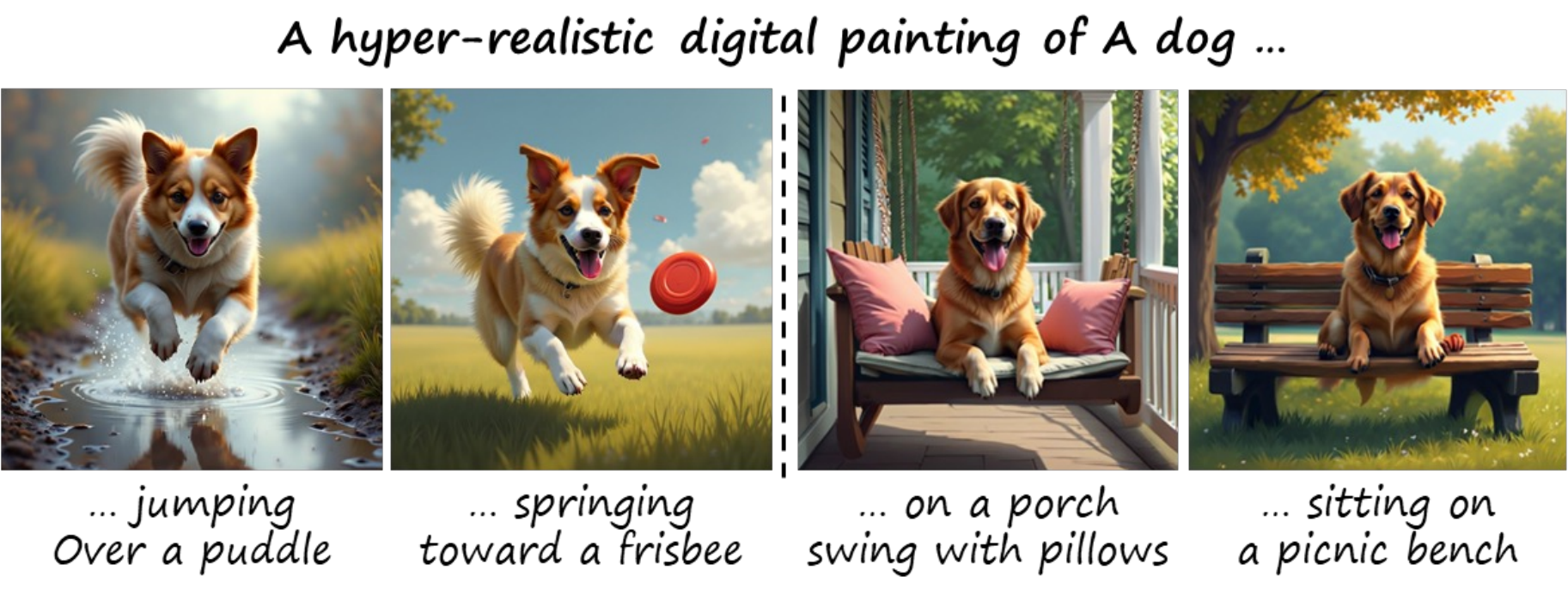}
    \caption{Context-bias in text-to-image generation.}
    \label{fig:bias}
\end{figure}

\subsection{Identity Prompt Replacement}
\label{sec:identity_prompt}

It is well known that generative models reflect biases in their training data distributions~\cite{zhou2024bias, wei2025addressing}. For instance, the prompt ``a dog springing toward a frisbee'' (dynamic expression) often generates a Welsh corgi, while ``a dog on a porch swing with pillows'' (static expression) tends to produce calm, domesticated dogs like a Golden retriever, as illustrated in \figref{fig:bias}—highlighting how prompt phrasing shapes semantic interpretation.
This bias stems from the self-attention mechanism, where identity representations (e.g., “a dog”) are influenced by their surrounding context, leading to inconsistent identity attributes---including gender, age, and species---across prompts.

To address the context bias inherent in text encoders, we propose an \textit{Identity Prompt Replacement} (IPR) strategy that reduces such bias through the alignment of identity-related attributes across prompts. Specifically, we enforce a consistent representation of identity by replacing all identity embeddings $\mathbf{T}_{\text{iden}} = \{T_{\text{iden}}^n\}_{n=1}^{N}$ with those extracted from a reference instance (by default, the first sample in the batch).
To maintain the proportional relationship between identity and expression features, we further normalize the magnitude of the expression embeddings $\mathbf{T}_{\text{exp}} = \{T_{\text{exp}}^n\}_{n=1}^{N}$ as follows:

\begin{equation}
\begin{gathered}
\mathbf{\hat{T}} = \left( \hat{\mathbf{T}}_{\text{iden}}, \hat{\mathbf{T}}_{\text{exp}} \right)
= \left\{ \left( T_{\text{iden}}^1, \frac{ \left\| T_{\text{iden}}^1 \right\| }{ \left\| T_{\text{iden}}^n \right\| } \cdot T_{\text{exp}}^n \right) \right\}_{n=1}^N,\\
\end{gathered}
\end{equation}
where $\hat{\mathbf{T}}_{\text{iden}}$ and $\hat{\mathbf{T}}_{\text{exp}}$ denote identity and expression prompt embeddings processed via IPR.

\subsection{Unified Attention Guidance}

\paragraph{Adaptive Style Injection}
\label{sec:adaptive}
Although the Identity Prompt Replacement (IPR) technique mitigates context-level discrepancies by aligning identity-related attributes across prompts, it remains insufficient in preserving the appearance-level identity and global visual style consistency.
To address this, we propose an \textit{Adaptive Style Injection} (ASI) mechanism that aligns both the appearance of the identity and the overall scene style. ASI operates within the self-attention layers during the early generation steps, motivated by prior findings that analyze the functional roles of generation stages for style alignment~\cite{park2025training}.

As illustrated in \figref{fig:overall}-(Right), for each sample in the batch, we replace all key features in the self-attention with those of the reference, i.e., $K_s^1$, which encourages the model to attend to semantically consistent regions anchored by the reference. Also, we compute the cosine similarity between its value features and those of a reference instance, to obtain an adaptive interpolation weight $\alpha_s^n$, which is then used to interpolate the value features, facilitating appearance-level alignment of identity. The Adaptive Style Injection is defined as follows:

\begin{equation}
\begin{gathered}
\bar{K}_s^n = K_s^{1}, 
\bar{V}_s^n = \alpha_s^n V_s^{n} + (1 - \alpha_s^n) V_s^{1}, \\
\alpha_s^n = \lambda \cdot \text{sim}(V_s^1, V_s^n),\forall n \in \{1, \cdots ,N\}, \\   
\label{eq:ASI}
\end{gathered}
\end{equation}
where $K_s^{1}$ and $V_s^{1}$ denote the key and value features of the reference sample, $V_s^{n}$ is the value feature of the $n$-th sample at $s$-th generation step, and $\lambda$ is a scaling coefficient. This similarity-guided adaptive operation facilitates the smooth and proportional alignment of appearance-level identity and global visual style across the batch, guided by the reference instance.

\paragraph{Synchronized Guidance Adaptation}
While Adaptive Style Injection improves identity appearance and global style consistency, applying it only to the conditional branch disrupts the balance between the conditional and unconditional signals established by Classifier-Free Guidance (CFG) \cite{ho2021classifierfree},  which is intended to enhance prompt fidelity. Such disruption may undermine the effectiveness of CFG, potentially degrading prompt fidelity in the generated images.

To resolve this, we propose \textit{Synchronized Guidance Adaptation}, which applies the same operation to the unconditional branch using the identical interpolation weights computed from the conditional path, as illustrated in \figref{fig:overall}-(Right). Specifically, for the unconditional branch, we modify the key and value features as:
\begin{equation}
\tilde{k}_s^n = k_s^{1}, \quad
\tilde{v}_s^n = \alpha_s^n v_s^{n} + (1 - \alpha_s^n) v_s^{1}, ~\forall n \in \{1,\cdots, N\},
\end{equation}
where $k_s^{1}$ and $v_s^{1}$ denote the unconditional branch’s key and value features of the reference sample, $v_s^{n}$ is the value feature of the $n$-th sample at $s$-th generation step, and $\alpha_s^n$ is the adaptive weight shared from the conditional branch \equref{eq:ASI}. By synchronizing the feature adaptation across both branches, our approach preserves the intended effect of classifier-free guidance, enabling generated images to faithfully reflect their text prompts while maintaining a consistent subject identity and overall style.

\section{Experiment}
\label{exp}

\subsection{Implementation details}
In our experiment, we leverage the pre-trained Infinity 2B model~\cite{han2024infinity} as our baseline. The model performs scale-wise prediction over 12 steps and employs a codebook with a dimensionality of $2^{32}$, producing quantized feature maps of resolution $64\times64$ with 32 channels. The early-stage step for Adaptive Style Injection and Synchronized Guidance Adaptation is fixed at $\mathbf{S}_{\text{early}}=\{2,3\}$, and the scaling coefficient $\lambda$ is set to 0.85. All other components of the model architecture remain unchanged, and all parameters are frozen throughout inference.

The number of output images is determined by the number of input text prompts. When generating four $1024\times1024$ images in parallel on a single A6000 GPU, the process takes approximately 6.88 seconds in total, or 1.72 seconds per image. For scenarios involving more than four prompts, we adopt a batched generation strategy: in each batch, the identity prompt paired with the first expression prompt is always placed first, while the remaining positions are filled with the other prompts. This approach ensures that identity information remains consistent and is effectively propagated across all generated batches.

\begin{table*}[t]

    \centering
    \resizebox{.9\textwidth}{!}{%
    \begin{tabular}{l|c|c|cccc|c}
    \toprule
     Method & Train-Free & $S_H$ $\uparrow$ & DINO $\uparrow$ & CLIP-T $\uparrow$ & CLIP-I $\uparrow$ & DreamSim $\downarrow$ & Inference Time (s) $\downarrow$ \\
    \midrule
     Vanilla SDXL \cite{podell2023sdxl} & - & 0.7408 & 0.6067 & 0.9074 & 0.8793 & 0.3385 & 10.27 \\
     Vanilla Infinity \cite{han2024infinity} & - & 0.7891 & 0.6965 & 0.8836 & 0.8955 & 0.2780 & 1.71 \\
    \midrule
     % Textual Inversion \cite{gal2022image} & \tikzxmark & 0.5465 & 0.3910 & 0.7340 & 0.7702 & 0.5250 & 19.52 \\
     IP-Adapter \cite{ye2023ipadaptertextcompatibleimage} & \tikzxmark & 0.8323 & \underline{0.7834} & 0.8661 & \underline{0.9243} & 0.2266 & \underline{10.40} \\
     PhotoMaker \cite{li2024photomaker} & \tikzxmark & 0.7223 & 0.6516 & 0.8651 & 0.8465 & 0.3996 & 19.52 \\
     The Chosen One \cite{avrahami2024chosen} & \tikzxmark & 0.6494 & 0.5824 & 0.8162 & 0.7943 & 0.4893 & 13.47\\
     OneActor \cite{wang2024oneactor} & \tikzxmark & 0.8088 & 0.7172 & 0.8859 & 0.9070 & 0.2423 & 24.94 \\
    \midrule
     ConsiStory \cite{tewel2024training} & \tikzcmark & 0.7902 & 0.6895 & \textbf{0.9019} & 0.8954 & 0.2787 & 37.76 \\
     StoryDiffusion \cite{zhou2024storydiffusion} & \tikzcmark & 0.7634 & 0.6783 & 0.8403 & 0.8917 & 0.3212 & 23.68 \\
     %1Prompt1Story \cite{liu2025onepromptonestory} & \tikzcmark & 0.8377 & 0.7687 & 0.8855 & \underline{0.9207} & \underline{0.2053} & 22.57 \\
     1Prompt1Story \cite{liu2025onepromptonestory} & \tikzcmark & \underline{0.8395} & 0.7687 & \underline{0.8942} & 0.9117 & \underline{0.1993} & 22.57 \\
     Ours & \tikzcmark & \textbf{0.8538} & \textbf{0.8089} & {0.8732} & \textbf{0.9267} & \textbf{0.1834} & \textbf{1.72} \\
    \bottomrule
    \end{tabular}
    }
    \caption{Quantitative comparison with state-of-the-art consistent T2I generation models. Inference time is reported per image. Symbols $\uparrow$ and $\downarrow$ indicate whether higher or lower values are better. \textbf{Bold} and \underline{underline} denote the best and second-best results, respectively.}
    \label{tab:quantitative}
\end{table*}

\subsection{Evaluation Setup}

\noindent\textbf{Benchmark.} We follow the evaluation protocol proposed in 1Prompt1Story \cite{liu2025onepromptonestory}, an extension of the original ConsiStory benchmark \cite{tewel2024training}. ConsiStory+ expands the evaluation space by introducing a more diverse range of subjects, prompt descriptions, and styles. In accordance with this setup, we evaluate both prompt alignment as well as the consistency of subject identity and style over 200 distinct prompt sets, resulting in the generation of up to 1,500 images in total.

\noindent\textbf{Evaluation Metrics.} Following 1Prompt1Story, to assess \textit{prompt fidelity}, we compute the average CLIP text score \cite{radford2021learning} between each generated image and its corresponding prompt, denoted as CLIP-T. For \textit{identity consistency}, we utilize DreamSim \cite{fu2023dreamsim}, a perceptual similarity metric shown to correlate well with human judgment, as well as CLIP-I \cite{radford2021learning}, which measures the cosine similarity between image embeddings. Following the protocol of DreamSim, we remove image backgrounds using CarveKit \cite{CarveKit} and replace them with random noise, so that the similarity measurements reflect only the subject’s identity. The same background removal process is applied to images evaluated with CLIP-I to ensure consistency across identity-based metrics. To assess \textit{style consistency} among images conditioned on the same identity prompt, we follow prior work on style-aligned image generation \cite{hertz2024style, park2025training, frenkel2024implicit} and compute the average pairwise DINO similarity, which captures alignment in overall visual appearance.
For a more comprehensive evaluation, we further report a harmonic score $S_H$, which aggregates four core metrics (CLIP-T, CLIP-I, DreamSim, and DINO) using the harmonic mean. Since DreamSim is a distance-based metric, we convert it to a similarity measure via [$1 - \text{DreamSim}$] before computing the mean. This composite score provides a balanced view of both prompt fidelity and visual consistency across identity and style.

\subsection{Comparison with state-of-the-art consistent text-to-image generation models}

\noindent\textbf{Quantitative comparison.}
\tabref{tab:quantitative} provides a comparative analysis of our method against a variety of state-of-the-art consistent text-to-image generation models, encompassing both training-based and training-free approaches. Despite requiring neither fine-tuning nor training, our method achieves the best $S_H$ score, reflecting a strong balance across all core metrics. Notably, our method achieves the highest DINO similarity, as well as leading scores in CLIP-I and DreamSim, demonstrating superior consistency in both style and identity. The identity metrics are computed after background removal to isolate subject appearance, further validating our approach's robustness in preserving subject identity across generated images. While 1Prompt1Story also delivers competitive results as a training-free baseline, our method surpasses it in both style and identity consistency, and overall $S_H$, while operating over 13$\times$ faster. Importantly, these results are achieved with an inference time of just 1.72 seconds per image—significantly faster than most diffusion-based models, which typically exceed 10 seconds. These results underscore that Infinite-Story not only provides high-quality and consistent generations but also does so with remarkable efficiency, making it suitable for practical deployment in real-time generation scenarios.

\begin{figure}[t]
    \centering
    \includegraphics[width=.9\linewidth]{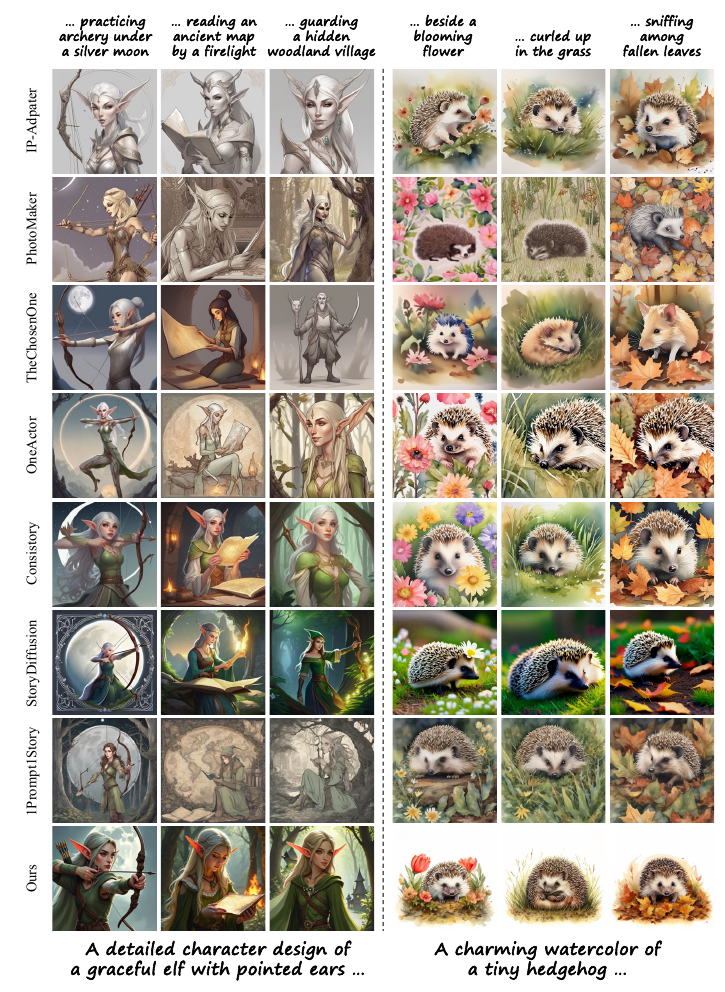}
    \caption{Qualitative comparison with state-of-the-art consistent T2I generation models. Each row depicts a set of images generated using a shared identity prompt combined with varying expression prompts.}
    \label{fig:qualitative}
\end{figure}

\begin{table}[t]
    \centering
    \resizebox{.47\textwidth}{!}{%

    \small
    \begin{tabular}{c|c c c| c| c c c c}
    \toprule

      \multicolumn{1}{c}{} & \multicolumn{3}{c}{Component} & \multicolumn{5}{c}{Quantitative Metrics} \\
      \# & IPR & ASI & SGA & $S_H$ $\uparrow$ & DINO $\uparrow$ & CLIP-T $\uparrow$ & CLIP-I $\uparrow$ & DreamSim $\downarrow$ \\
    \midrule
     (a) &  &  &  & 0.7891 & 0.6965 & \textbf{0.8836} & 0.8955 & 0.2780 \\
     (b) & \tikzcmark &  &  & 0.8013 & 0.7119 & \underline{0.8814} & 0.9046 & 0.2569 \\
     (c) & \tikzcmark & \tikzcmark &  & \underline{0.8481} & \underline{0.8082} & 0.8625 & \underline{0.9242} & \underline{0.1931} \\
     (d) & \tikzcmark & \tikzcmark & \tikzcmark & \textbf{0.8538} & \textbf{0.8089} & {0.8732} & \textbf{0.9267} & \textbf{0.1834}  \\
    \bottomrule
    \end{tabular} 
    }
    \caption{Ablation study on the Identity Prompt Replacement (IPR), Adaptive Style Injection (ASI), and Synchronized Guidance Adaptation (SGA). The symbol $\uparrow$ indicates that higher values are better, and $\downarrow$ indicates that lower values are better. The best and second-best results are highlighted in \textbf{bold} and \underline{underline}, respectively.}
    \label{tab:ablation}
\end{table}

\begin{figure}[h]
  \centering
  \includegraphics[width=.9\linewidth]{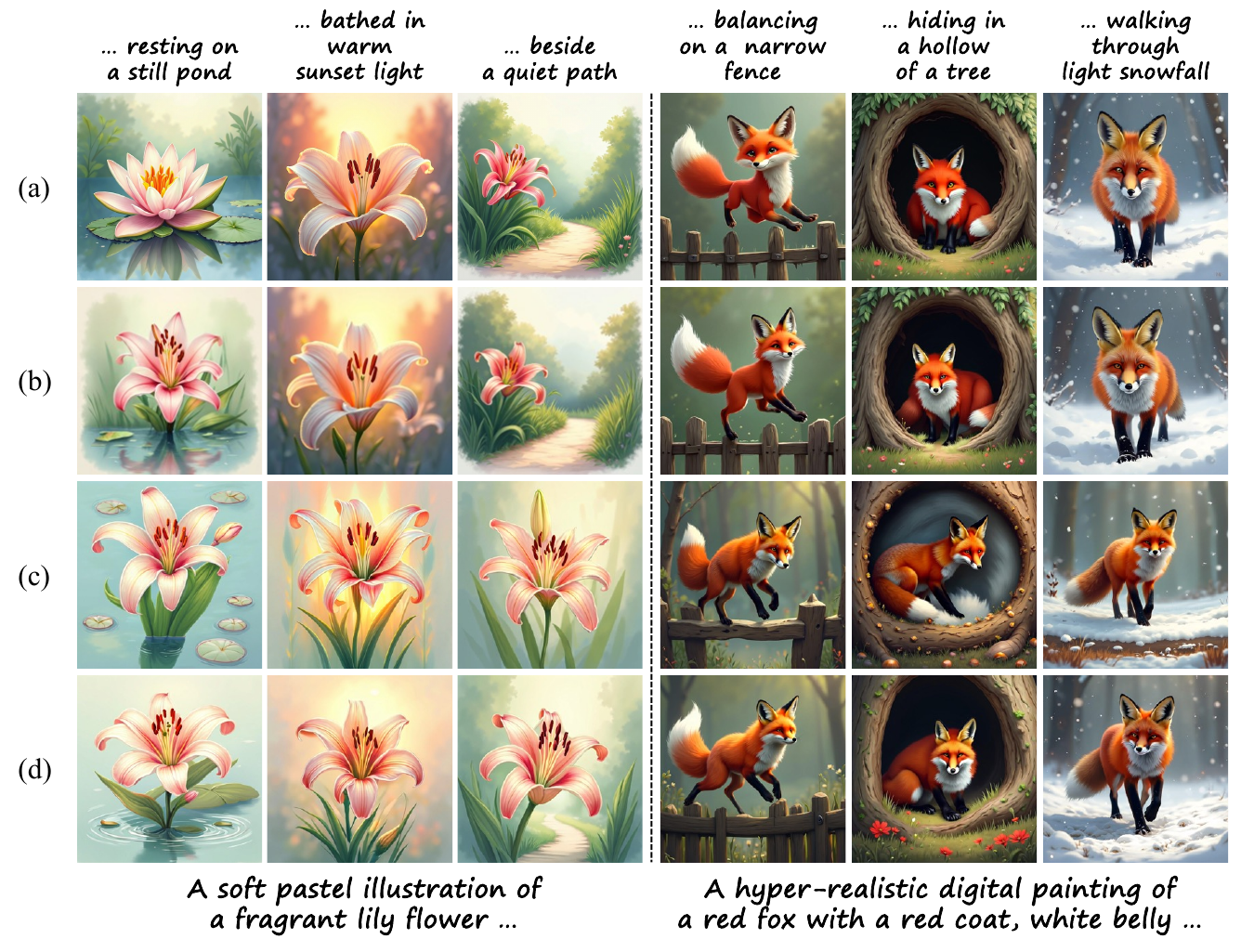}
  \caption{Qualitative analysis of ablation study. The results from (a)-(d) correspond to the configurations in \tabref{tab:ablation}.}
  \label{fig:ablation}
\end{figure}

\noindent\textbf{Qualitative comparison.} \figref{fig:qualitative} presents qualitative results across two themes—an elf character and a watercolor-style hedgehog—used to evaluate various consistent text-to-image generation models. 
% Each row depicts a set of images conditioned on a shared identity prompt and varying expression prompts. 
% We compare the ability of each model to preserve subject identity, maintain style consistency, and prompt fidelity.
Some methods, such as IP-Adapter, preserve subject identity effectively, especially in facial structure and posture. However, they often fail to reflect prompt-specific nuances, as seen in the elf example, where different expression settings (e.g., “guarding a hidden woodland village”) result in minimal contextual variation. On the other hand, OneActor and 1Prompt1Story capture expression prompts well but show style shifts in background and rendering details that disrupt visual cohesion.
StoryDiffusion and ConsiStory demonstrate style consistency, yet they exhibit inconsistencies in subject identity across the prompt set.
PhotoMaker and The Chosen One, while producing aesthetically pleasing results, tend to underperform across all three aspects—prompt fidelity, identity consistency, and style consistency.

Our model successfully addresses all three aspects. In both the elf and hedgehog scenarios, the generated images reflect clear variation across prompts while consistently preserving subject identity and maintaining a unified visual style. These results confirm that our method generates image sequences that are identity-consistent, style-consistent, and faithful to the prompt.

\subsection{Ablation study}
\noindent\textbf{Quantitative Analysis.}
\tabref{tab:ablation} presents an ablation study evaluating the contributions of each proposed component. Starting with \textit{Identity Prompt Replacement} (IPR), as shown in row (b), we observe notable gains in CLIP-I and DreamSim, confirming its effectiveness in aligning identity-related attributes across prompts by mitigating the context bias of text encoders.
When \textit{Adaptive Style Injection} (ASI) is added in (c), DINO similarity increases significantly, indicating improved global style consistency. Additionally, CLIP-I and DreamSim scores also improve, reflecting enhanced alignment in identity appearance. 
Finally, in (d), applying \textit{Synchronized Guidance Adaptation} (SGA) helps restore the balance between the conditional and unconditional branches of CFG, leads to a meaningful gain in CLIP-T, and further consolidates overall consistency.
Although there is a slight trade-off in prompt fidelity compared to the baseline, the full configuration achieves the highest harmonic score $S_H$, indicating that our method successfully balances identity coherence, style consistency, and prompt fidelity—all without any additional fine-tuning.

\noindent\textbf{Qualitative analysis.} 
\figref{fig:ablation} presents qualitative results from our ablation study on the proposed methods.
Without any proposed methods (a), the generated images exhibit severe inconsistency in both subject identity and visual style. For instance, the flower species and rendering styles vary across scenes, and the red fox’s appearance—such as fur texture and facial shape—fluctuates noticeably.
Introducing only \textit{Identity Prompt Replacement} (IPR) in (b) improves identity-related attributes by mitigating the context bias of text encoders. The lily maintains a more unified floral structure across prompts, and the red fox preserves more consistent facial features and body proportions. However, global style elements—such as lighting and rendering—remain inconsistent.
When \textit{Adaptive Style Injection} (ASI) is added (c), both global style and appearance-level identity consistency are further enhanced. The flower exhibits stable coloration and stroke patterns, while the red fox retains consistent shading and background textures across diverse scenes. Nevertheless, some prompt-specific semantics remain underemphasized, and visual artifacts such as unnatural outlines or distorted textures occasionally appear—likely due to strong style injection overriding localized details.
Finally, applying the full method with \textit{Synchronized Guidance Adaptation} (SGA) in (d) restores balance between the conditional and unconditional branches, enabling better preservation of prompt fidelity. This results in visually coherent outputs that maintain consistent subject appearance and unified stylistic rendering, while accurately reflecting prompt-specific variations—evidenced by appropriate posture, context, and lighting across prompts. 
These qualitative trends are consistent with the quantitative improvements observed in \tabref{tab:ablation}.

\begin{table}[t]
\centering
\resizebox{0.45\textwidth}{!}{%
\footnotesize
\begin{tabular}{l|ccc}
\toprule
Method & Identity $\uparrow$ &Style $\uparrow$& Prompt $\uparrow$ \\
\midrule
1Prompt1Story \cite{liu2025onepromptonestory} & 18.0\% & 13.2\% & 28.2\% \\
OneActor \cite{wang2024oneactor} & 7.2\% & 7.2\% & 10.6\% \\
IP-Adapter \cite{ye2023ipadaptertextcompatibleimage} & 16.4\% & 29.6\% & 4.7\% \\
\midrule
\textbf{Ours} & \textbf{58.4\%} & \textbf{50.0\%} & \textbf{56.5\%} \\
\bottomrule
\end{tabular}
}
\caption{User study preference percentages.}

\label{tab:user_study}
\end{table}

\subsection{User study} 
To complement our quantitative evaluation, we conduct a user study, with results shown in \tabref{tab:user_study}. A total of 55 participants were asked to assess three core criteria: identity consistency, prompt fidelity, and style consistency. We compare images generated by our model with those from 1Prompt1Story \cite{liu2025onepromptonestory}, OneActor \cite{wang2024oneactor}, and IP-Adapter \cite{ye2023ipadaptertextcompatibleimage}, which ranked highest in our quantitative benchmarks. The results of the user study indicate that our model consistently outperforms competing methods in all three aspects—identity, prompt, and style consistency—demonstrating strong human-perceived performance across a variety of prompts.

\section{Conclusion}
\label{conclusion}
In this paper, we present \textit{Infinite-Story}, a training-free framework for consistent text-to-image generation tailored to multi-prompt scenarios. Built upon a scale-wise autoregressive backbone, our method tackles two key challenges in consistent generation—identity inconsistency and style inconsistency—without requiring model fine-tuning or training. To this end, we introduce three lightweight yet effective techniques: \textit{Identity Prompt Replacement}, which mitigates the context bias of text encoders to align identity-related attributes, and a unified attention guidance strategy that combines \textit{Adaptive Style Injection} and \textit{Synchronized Guidance Adaptation} to align appearance-level identity and global style while preserving prompt fidelity. Extensive experiments show that Infinite-Story achieves state-of-the-art identity and style consistency while preserving diversity, and runs over 6$\times$ faster than leading diffusion models, making it practical for real-time and interactive applications such as visual storytelling.
Future work includes extending our method to support temporal consistency in video generation and exploring more adaptive reference selection strategies.

\section*{Acknowledgments}
This research was supported by the 2024 innovation base artificial intelligence data convergence project project with the funding of the 2024 government (Ministry of Science and ICT) (S2201-24-1002), Institute of Information \& Communications Technology Planning \& Evaluation(IITP) grant funded by the Korea government(MSIT) (No. RS-2025-02219277, AI Star Fellowship Support Project(DGIST)), Basic Science Research Program through the National Research Foundation of Korea (NRF) funded by the Ministry of Education (RS-2025-25420118) and LG AI STAR Talent Development Program for Leading Large-Scale Generative AI Models in the Physical AI Domain (RS-2025-25442149).

\bibliography{aaai2026}

% \clearpage
% \input{ReproducibilityChecklist/LaTeX/ReproducibilityChecklist}

\clearpage
\appendix
\begin{center}
    \LARGE \textbf{Infinite-Story: A Training-Free Consistent Text-to-Image Generation Appendix}
\end{center}

\section{Evaluation Details}
\label{sec:evaluation}
\subsection{Evaluation protocol}
To ensure consistency and comparability across models, we follow the evaluation protocol established by 1Prompt1Story~\cite{liu2025onepromptonestory}, with an additional criterion of \textit{style consistency}. The evaluation covers three key aspects: \textit{prompt fidelity}, \textit{identity consistency}, and \textit{style consistency}.

\paragraph{Prompt Fidelity (CLIP-T).}
To assess \textit{prompt fidelity}, we compute the CLIP text similarity (CLIP-T) using CLIP ViT-B/32~\cite{radford2021learning}. Following~\cite{liu2025onepromptonestory}, we prepend the prefix \textit{``A photo depicts''} to each prompt and scale the cosine similarity by a factor of 2.5. The CLIP-T score is computed between the generated image and its paired prompt, and the final score is obtained by averaging across all samples.

\paragraph{Identity Consistency (CLIP-I and DreamSim).}
We measure \textit{identity consistency} using two metrics: CLIP image similarity (CLIP-I) and DreamSim~\cite{fu2023dreamsim}. Both are computed as the average pairwise similarity between images generated from the same identity prompt. To remove background bias, we apply CarveKit~\cite{CarveKit} to extract the foreground subject and replace the background with uniform random noise, as done in~\cite{liu2025onepromptonestory}. CLIP-I uses the ViT-B/16 model, and DreamSim provides perceptual similarity scores aligned with human judgment.

\paragraph{Style Consistency (DINO).}
To assess \textit{style consistency} among images conditioned on the same identity prompt, we follow prior work on style-aligned image generation~\cite{hertz2024style, park2025training, frenkel2024implicit} and compute the average pairwise DINO similarity. Specifically, we use the CLS token from the DINO ViT-B/8~\cite{caron2021emerging} model to measure global visual similarity, capturing alignment in rendering, background, and texture.

\paragraph{Harmonic Score ($S_H$).}
To jointly evaluate consistency across all dimensions, we report a harmonic score $S_H$ defined as:

\begin{equation}
    S_H = \text{HM}(\text{CLIP-T}, \text{CLIP-I},1-\text{DreamSim}, \text{DINO}),
\end{equation}
where HM denotes Harmonic Mean. This combined metric penalizes inconsistency in any single component, providing a robust measure of overall generation quality.

\paragraph{Implementation.}
We adopt the official evaluation scripts from~\cite{liu2025onepromptonestory}, with minor adaptations (Add DINO metric). All metrics are computed on a single A6000 GPU using PyTorch. Background removal is applied consistently for identity-based metrics, and all features are extracted following standard preprocessing pipelines provided by each model. 

\subsection{Image-based consistent text-to-image models}
\label{app:imagebased}

We compare our method against several image-based consistent text-to-image generation approaches that leverage Stable Diffusion XL \cite{podell2023sdxl} as their backbone. Specifically, we include the following representative models:

\begin{itemize}

    \item \textbf{IP-Adapter}~\cite{ye2023ipadaptertextcompatibleimage}: We use the official code.
    \item \textbf{PhotoMaker}~\cite{li2024photomaker}: We use the official code.
    \item \textbf{StoryDiffusion}~\cite{zhou2024storydiffusion}: We use the official repository.
    \item \textbf{OneActor}~\cite{wang2024oneactor}: We use the official repository.
\end{itemize}

For all methods, we adopt the default DDIM sampling settings provided in their open-source implementations. During inference, each of these models requires an external reference image as an additional input (one or more). Therefore, we generate the reference image by providing only the identity portion of the full prompt to the corresponding base model. For instance, given the prompt ``A graceful unicorn galloping through a flower field,'' we generate the reference image using ``A graceful unicorn.'' This image is then used consistently across all prompts in the same sequence.

\subsection{Non-reference consistent text-to-image models}
\label{app:others}
We also compare our method against non-reference consistent text-to-image generation approaches that leverage Stable Diffusion XL \cite{podell2023sdxl} as their backbone. Specifically, we include the following representative models:

\begin{itemize}
    \item \textbf{The chosen one}~\cite{ye2023ipadaptertextcompatibleimage}: We use the unofficial code.
    \item \textbf{ConsiStory}~\cite{tewel2024training}: We use the official code.
    \item \textbf{1Prompt1Story}~\cite{wang2024oneactor}: We use the official repository.
\end{itemize}

For all methods, we adopt the default DDIM sampling settings provided in their open-source implementations. To ensure consistency across comparisons, we fix the number of DDIM sampling steps to 50 for all models, including the unofficial implementation of The Chosen One.

\begin{table*}[t]
    \centering
    \resizebox{0.8\linewidth}{!}{%
    \begin{tabular}{l|c|cccc}
    \toprule
     Method & $S_H$ $\uparrow$ & DINO $\uparrow$ & CLIP-T $\uparrow$ & CLIP-I $\uparrow$ & DreamSim $\downarrow$ \\
    \midrule

     Vanilla Switti \cite{voronov2024swittidesigningscalewisetransformers} & 0.7719 & 0.6595 & \textbf{0.8904} & 0.8871 & 0.2934 \\
     Switti + Ours & \textbf{0.8146} & \textbf{0.7441} & 0.8756 & \textbf{0.9018} & \textbf{0.2398} \\
    \midrule
     Vanilla HART \cite{tang2024hartefficientvisualgeneration} & 0.7434 & 0.6381 & \textbf{0.8848} & 0.8714 &  0.3488 \\
     HART + Ours & \textbf{0.7894} & \textbf{0.7048} & 0.8505 & \textbf{0.8982} & \textbf{0.2945} \\
    \bottomrule
    \end{tabular}
    }
    \caption{Effect of our technique on other scale-wise autoregressive model family.}
    \label{tab:othermodel}
\end{table*}

\subsection{Details of User Study}
To complement our quantitative evaluation, we conducted a user study with 50 participants, aged between 20 and 50. Each participant was shown a prompt along with four sets of generated images, each corresponding to a different method: our Infinite-Story, 1Prompt1Story~\cite{liu2025onepromptonestory}, OneActor~\cite{wang2024oneactor}, and IP-Adapter~\cite{ye2023ipadaptertextcompatibleimage}. 

Participants were instructed to select the image set that best satisfied each of the following criteria:

\begin{itemize}
    \item \textbf{Identity consistency}: Please select the option, from Option 1 to Option 4, that you find to have \underline{the most consistent appearance} of the subject.
    \item \textbf{Prompt fidelity}: Please select the option, from Option 1 to Option 4, that you find to have \underline{the most consistent style} throughout the image set.
    \item \textbf{Style consistency}: Please select the option, from Option 1 to Option 4, that you think \underline{best matches the text description}.
\end{itemize}

\begin{figure*}[h]
    \centering
    \includegraphics[width=1\linewidth]{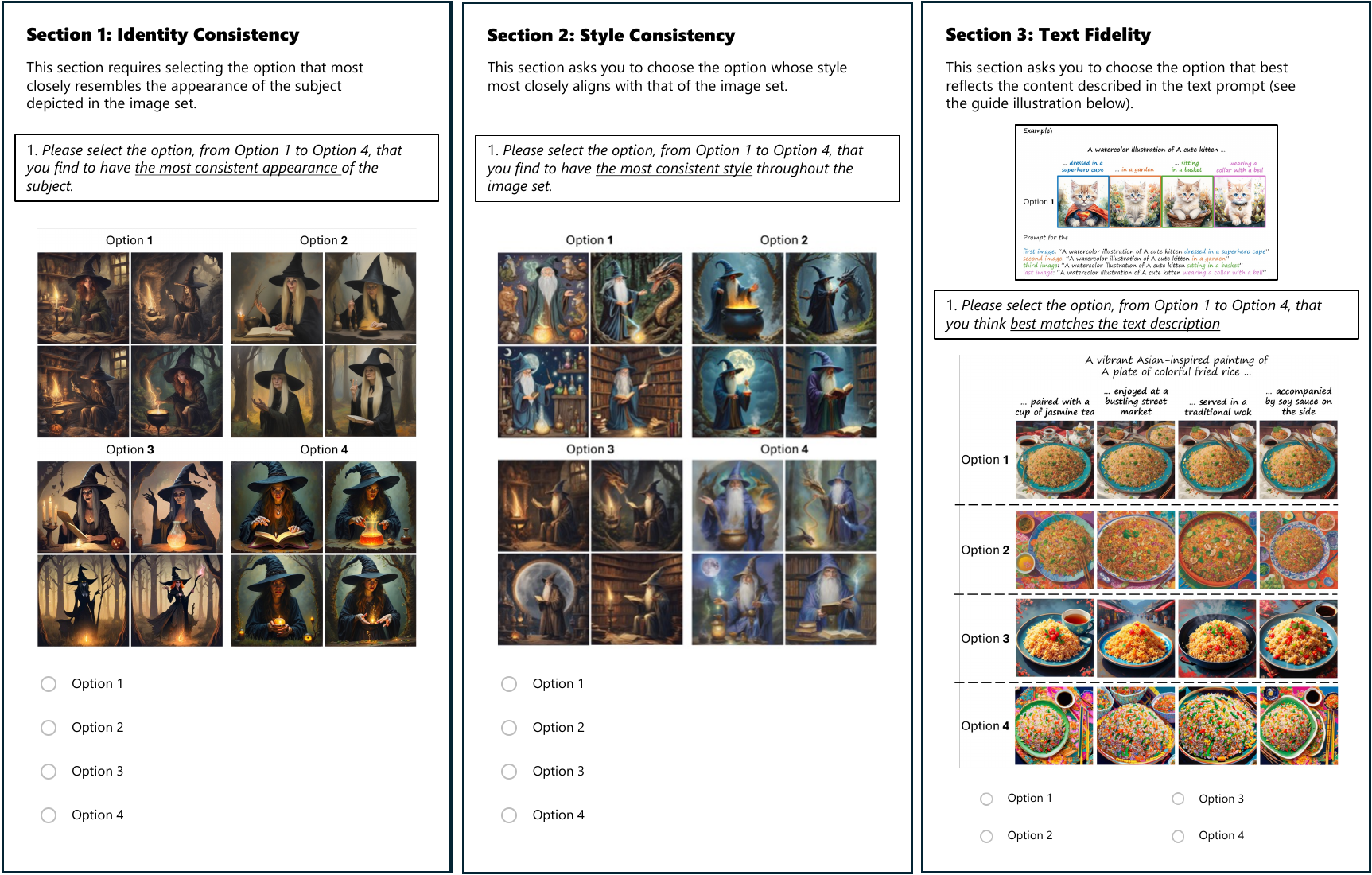}
    \caption{Example interface used in the user study. Participants selected the best-performing method among four candidates for each evaluation criterion.}
    \label{fig:userstudy_ui}
\end{figure*}

Each participant evaluated multiple sets across diverse prompts in randomized order. An example of the interface used in the study is shown in \figref{fig:userstudy_ui}.

\section{Comprehensive Analysis of Identity Prompt Replacement (IPR)}

\begin{figure}[h]
    \centering
    \includegraphics[width=1.0\linewidth]{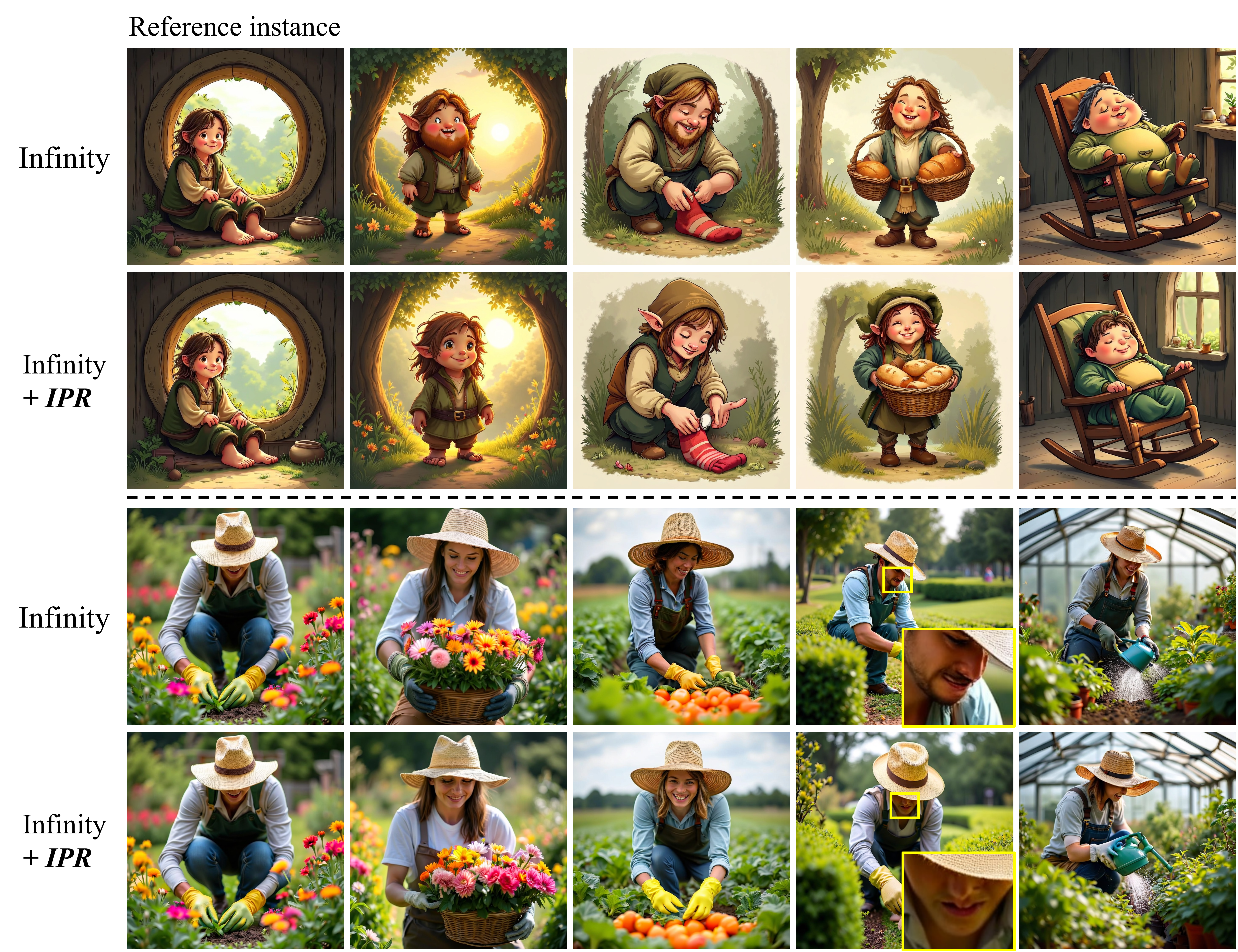}
    \caption{Qualitative analysis of Identity Prompt Replacement (IPR).}
    \label{fig:ablation_ipr}
\end{figure}

To further illustrate the effect of the Identity Prompt Replacement (IPR) module, we present a qualitative analysis of its influence on the generated images. Rather than copying the exact visual appearance of the object across different scenes, IPR helps preserve essential semantic attributes, such as gender and age, across all frames, as shown in \figref{fig:ablation_ipr}. This alignment of intrinsic characteristics plays a key role in enhancing the qualitative perception of coherence across the generated images. 
For instance, depending on the attributes of the reference instance, \figref{fig:ablation_ipr}-(top) shows that the baseline model tends to generate an older-looking object with a beard, whereas IPR adjusts the age to better match that of the reference. Similarly, \figref{fig:ablation_ipr} demonstrates that IPR aligns the gender of the generated object with the reference, which is especially evident in the yellow boxes in the fourth column.

\section{Generality of our method}
 To demonstrate the generalization capability, we applied our method to other scale-wise autoregressive text-to-image (T2I) generation models, Switti \cite{voronov2024swittidesigningscalewisetransformers} and HART \cite{tang2024hartefficientvisualgeneration}.
As shown in \tabref{tab:othermodel}, our method demonstrates clear performance improvements across both models, especially in DINO, CLIP-I, and DreamSim.
These results highlight the proposed technique's generalization capability beyond the infinity model and are applicable to a broader class of scale-wise autoregressive architectures.

\section{Additional Ablation study on Adaptive Style Injection}

We conduct an additional ablation study on the scaling coefficient $\lambda$ used in Adaptive Style Injection. This experiment aims to analyze how different values of $\lambda$ influence each evaluation metric. As shown in \tabref{tab:tau_ablation}, increasing $\lambda$ generally improves identity and style consistency metrics such as DINO, CLIP-I, and DreamSim. Notably, the best DINO and DreamSim scores are achieved when $\lambda$=0.9, indicating strong identity and style consistency. However, CLIP-T, which measures prompt fidelity, tends to degrade as $\lambda$ increases, achieving its highest value at $\lambda$ = 0.6. To strike a better balance between prompt fidelity and consistency, we select $\lambda$=0.85 as our default, which outperforms $\lambda$=0.9 in CLIP-T (0.8732 vs. 0.8722) while still maintaining competitive performance in other metrics.

\begin{table*}[t]
    \centering
    \begin{tabular}{l|c|cccc}
    \toprule
     Parameter & $S_H$ $\uparrow$ & DINO $\uparrow$ & CLIP-T $\uparrow$ & CLIP-I $\uparrow$ & DreamSim $\downarrow$ \\
    \midrule
     $\lambda$ = 0.6 & {0.8420} & {0.7967} & \textbf{0.8745} & {0.9209} & {0.1998} \\
     $\lambda$ = 0.7  & {0.8473} & {0.7865} & \underline{0.8737} & {0.9227} & {0.1919}  \\
      $\lambda$ = 0.8  & {0.8506} & {0.8058} & 0.8735 & {0.9245} & {0.1904}  \\
     $\lambda$ = 0.85 (Ours)  & \textbf{0.8538} & \underline{0.8089} & {0.8732} & \textbf{0.9267} & \underline{0.1834} \\
      $\lambda$ = 0.9  & \textbf{0.8538} & \textbf{0.8102} & 0.8722 & \underline{0.9251} & \textbf{0.1826}  \\
    \bottomrule
    \end{tabular}
    \caption{Ablation study on the Adaptive Style Injection scaling coefficient $\lambda$. Symbols $\uparrow$ and $\downarrow$ indicate whether higher or lower values are better. \textbf{Bold} and \underline{underline} denote the best and second-best results, respectively.}
    \label{tab:tau_ablation}
\end{table*}

\section{Limitation of Infinite-story}
Our Infinite-Story relies on a single reference image (anchor) within each batch to propagate identity and style features. While this enables efficient and training-free inference, it introduces sensitivity to anchor selection. If the anchor image is of low quality or stylistically off-target, this degradation may propagate to the entire batch. Notably, as our method does not alter the generation capabilities of the underlying Infinity model, its success is inherently tied to the quality of the initial output. This highlights the importance of future work in developing adaptive anchor selection or correction mechanisms.

\section{Long story generation}
\begin{figure*}[h]
    \centering
    \includegraphics[width=0.85\linewidth]{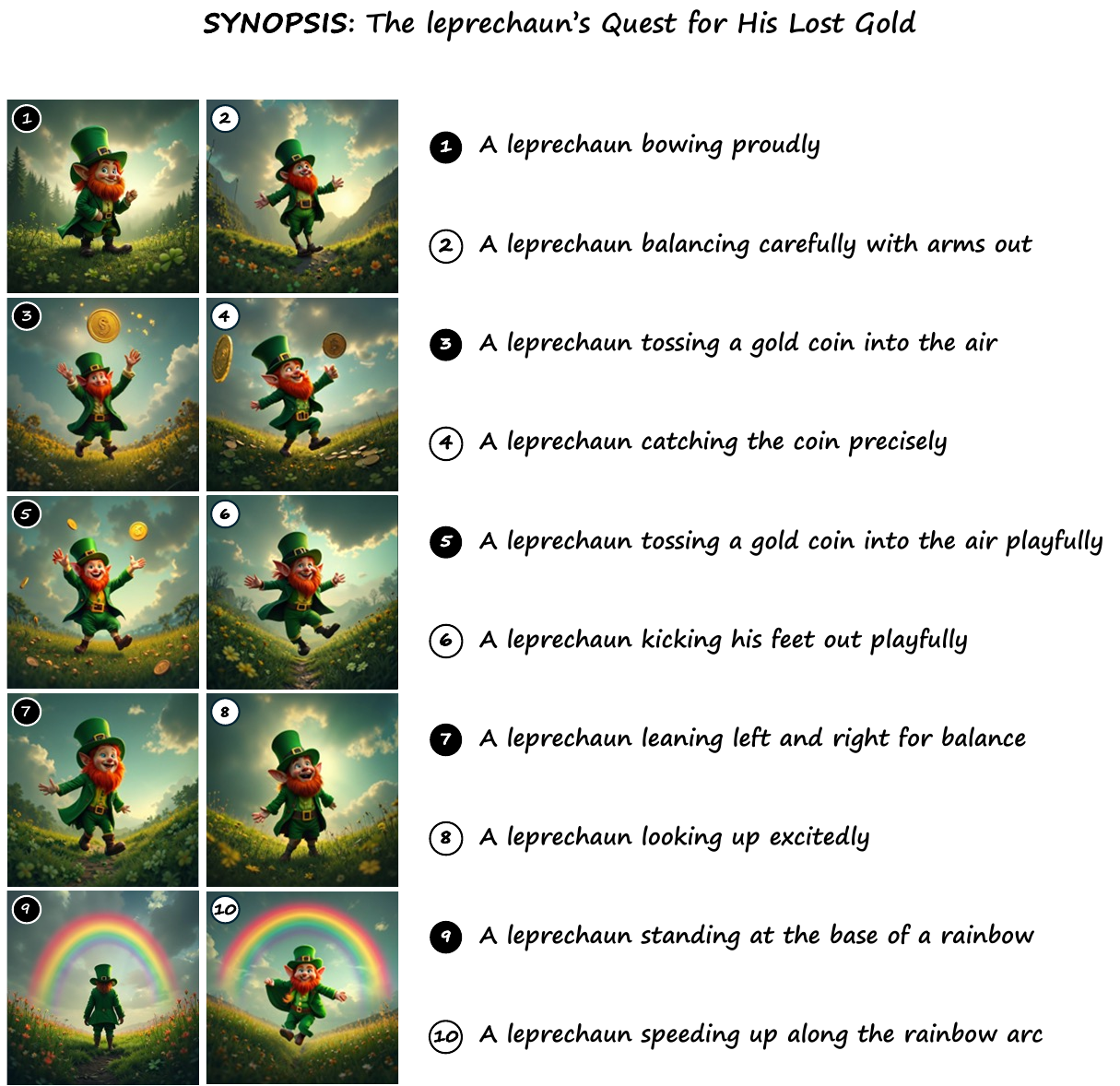}
    \caption{Long story generated by Infinite-Story. Under a unified synopsis, the leprechaun's quest for finding gold coin unfolds through organically connected scenes while preserving consistency of identity and style.}
    \label{fig:longstory}
\end{figure*}

\begin{figure*}[t]
    \centering
    \includegraphics[width=0.85\linewidth]{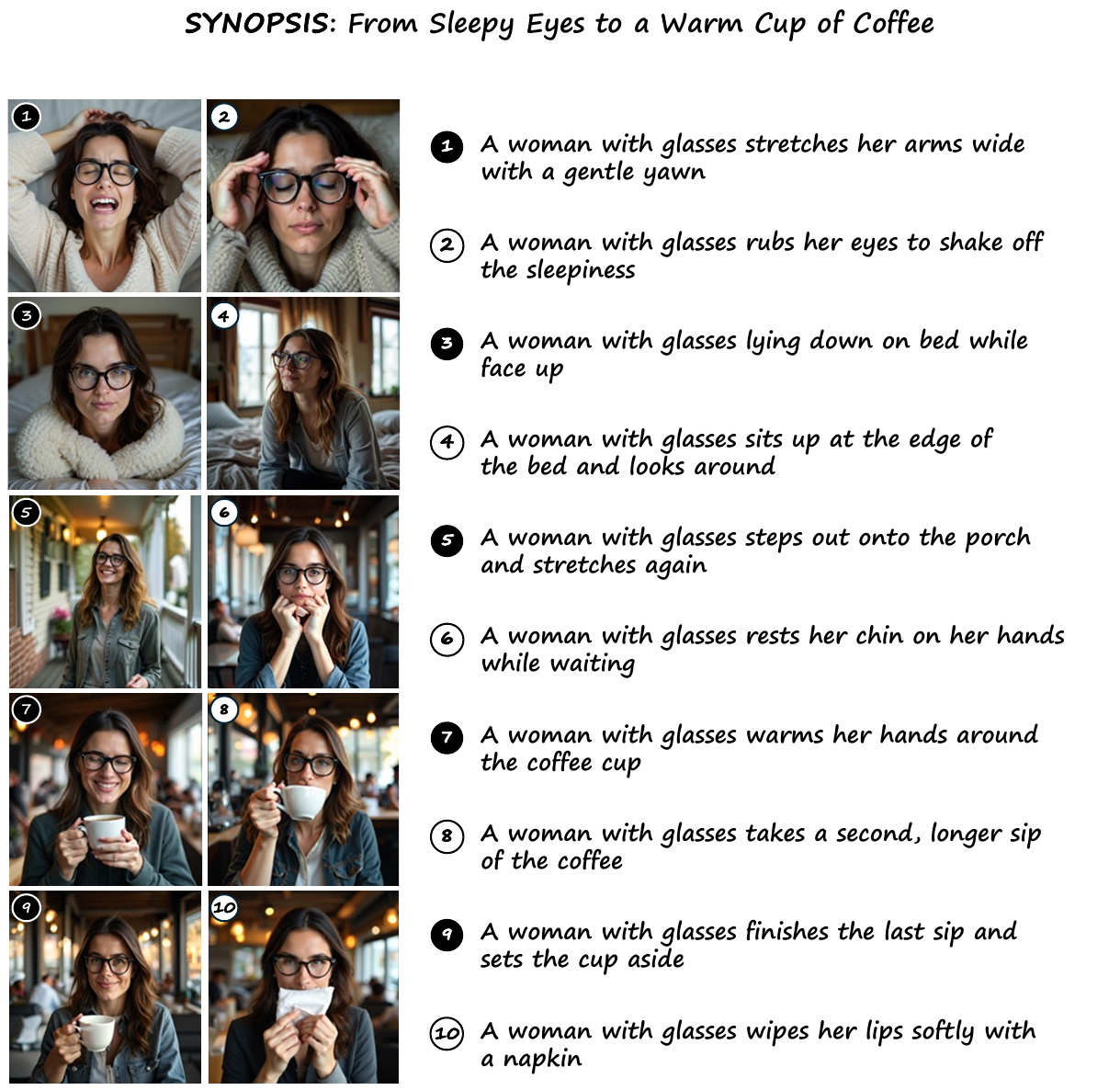}
    \caption{Long story generated by Infinite-Story. Under a unified synopsis, the story of woman with glasses transitions from waking up to enjoying her coffee at cafe unfolds through organically connected scenes while preserving consistency of identity and style.}
    \label{fig:longstory2}
\end{figure*}

To demonstrate the effectiveness of our method in generating extended, coherent visual narratives, we present two long-form examples in \figref{fig:longstory} and \figref{fig:longstory2}. Each figure illustrates a 10-frame story, where each image is conditioned on a unique prompt that reflects fine-grained scene variations while maintaining identity and style consistency.

\figref{fig:longstory} showcases a fantasy narrative titled \textit{“The leprechaun’s Quest for His Lost Gold”}, where a leprechaun performs a series of playful actions leading to the discovery of a rainbow. The consistent character appearance and stylistic rendering across dynamic scenes demonstrate the model’s ability to retain coherence in identity and global style throughout an extended sequence.

\figref{fig:longstory2} presents a slice-of-life story titled \textit{“From Sleepy Eyes to a Warm Cup of Coffee”}, following a woman’s transition from waking up to enjoying her coffee. Despite diverse poses, expressions, and indoor/outdoor transitions, the generated images preserve a unified visual identity and aesthetic style.

These results highlight Infinite-Story’s ability to support rich, multi-prompt storytelling applications such as comic strips, storyboarding, and animated content creation—all while preserving consistency across identity, style, and prompt fidelity.

\section{Addtional qualitative results}
\begin{figure*}[h]
    \centering
    \includegraphics[width=0.85\linewidth]{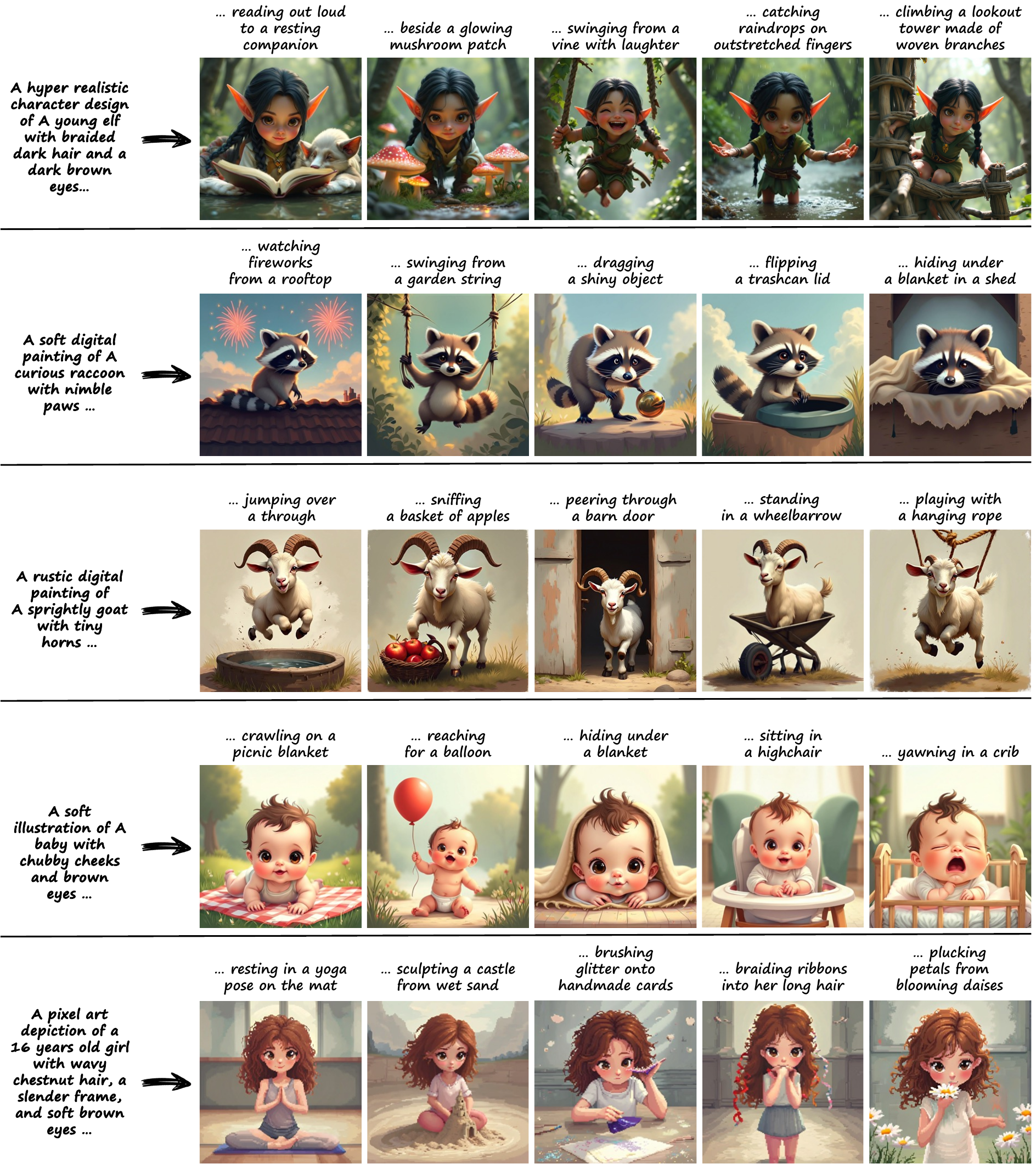}
    \caption{Additional qualitative results of Infinite-Story. Each row presents a set of images generated with a shared identity prompt combined with diverse expression prompts.}
    \label{fig:qual_results1}
\end{figure*}

\begin{figure*}[h]
    \centering
    \includegraphics[width=0.85\linewidth]{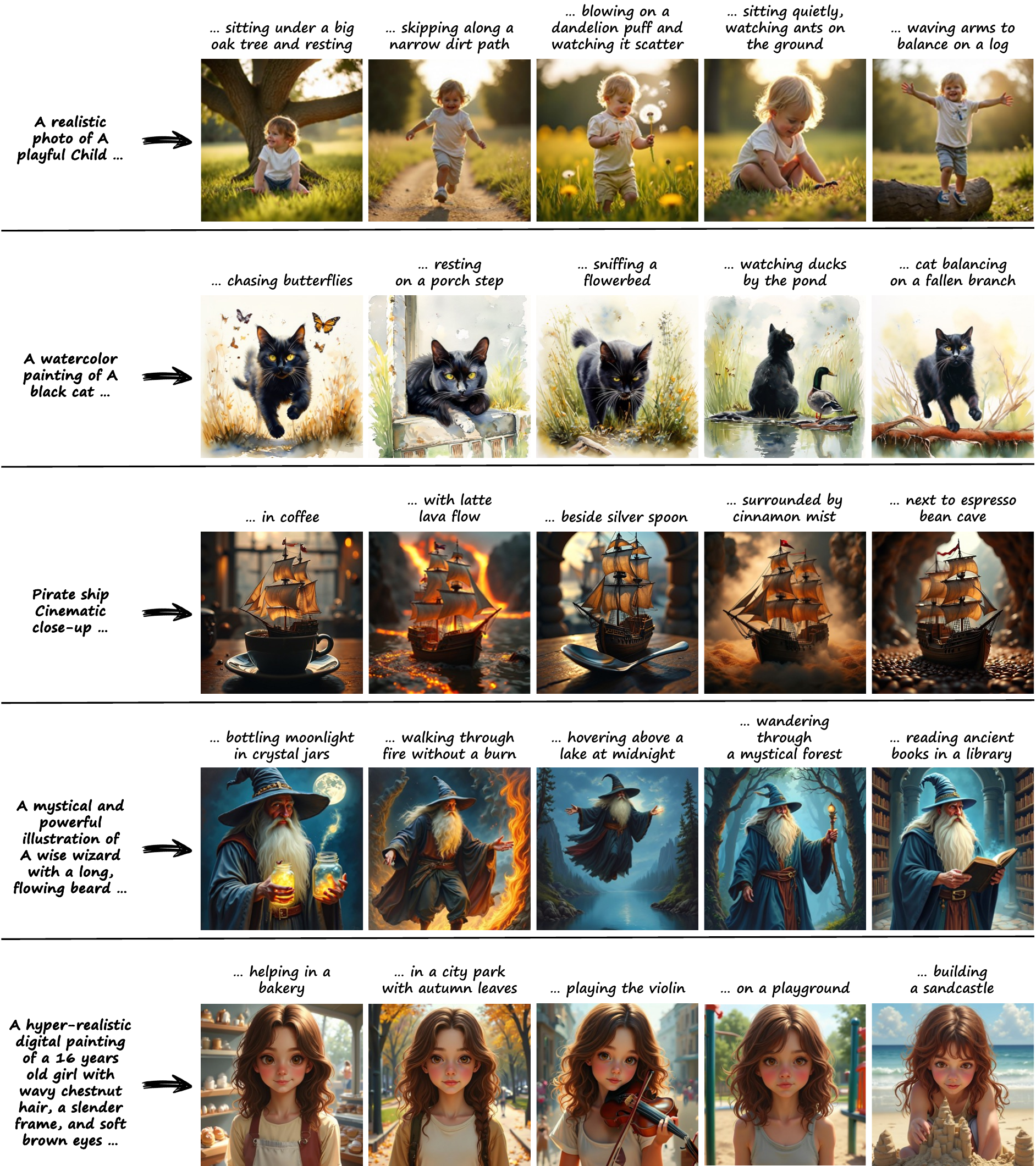}
    \caption{Additional qualitative results of Infinite-Story. Each row presents a set of images generated with a shared identity prompt combined with diverse expression prompts.}
    \label{fig:qual_results2}
\end{figure*}

We present additional qualitative results of our Infinite-Story in \figref{fig:qual_results1} and \figref{fig:qual_results2}. These additional results demonstrate that our method successfully preserves both identity and style consistency across diverse scenarios, while accurately reflecting the given text prompt. These results highlight the potential for broader practical applicability in various fields.

\end{document}